\pdfoutput=1
\documentclass[10pt,twocolumn,letterpaper]{article}

\usepackage[pagenumbers]{cvpr}

\usepackage[normalem]{ulem}

\usepackage{pifont}
\usepackage{xcolor}

\newcommand{\cmarkc}{\textcolor{green!60!black}{\ding{51}}}
\newcommand{\xmarkc}{\textcolor{red!70!black}{\ding{55}}}
\usepackage[T1]{fontenc}

\newcommand{\method}{SplatSeg3D\xspace}

\usepackage{tikz}
\usetikzlibrary{calc,arrows.meta}
\newcommand{\tmark}[1]{\tikz[remember picture,overlay]\coordinate(#1);}
\usepackage{algorithm}
\usepackage{algpseudocode}
\AtEndPreamble{\crefname{algorithm}{Alg.}{Algs.}}

\newcommand{\ours}{\textbf{\method (ours)}}

\newcommand{\est}[1]{#1}

\DeclareMathOperator*{\argmax}{arg\,max}

\usepackage{tabularx}
\usepackage{multirow}
\usepackage{booktabs}
\usepackage{amsmath}
\usepackage{amssymb}

\definecolor{cvprblue}{rgb}{0.21,0.49,0.74}
\usepackage[pagebackref,breaklinks,colorlinks,allcolors=cvprblue]{hyperref}

\usepackage{microtype}

\def\confName{CVPR}
\def\confYear{2027}

\title{SplatSeg3D: Efficient Zero-Shot 3D Segmentation \\ via Point Cloud Surfelization and Splatting}

\author{Sandeep Inuganti$^{*1,2}$ \and
Mahdi Chamseddine$^{*1}$ \and
Marcel Rogge$^{1,2}$ \and Martin R. Oswald$^{3}$ \and Didier Stricker$^{1,2}$ \and Jason Rambach$^{1}$ \and
$^{1}${\small German Research Center for Artificial Intelligence, DFKI, Germany}\\ 
$^{2}${\small RPTU Kaiserslautern, Germany} $^{3}${\small University of St. Gallen, Switzerland}\\
$^{1}${\tt\small \{firstname.lastname\}@dfki.de} $^{3}${\tt\small \{firstname.lastname\}@unisg.ch}}

\begin{document}
\maketitle
\makeatletter
\def\blfootnote{\xdef\@thefnmark{}\@footnotetext}
\makeatother
\blfootnote{*equal contribution}
\begin{abstract}
Open-vocabulary 3D segmentation methods rely on 2D foundation models applied to registered images captured alongside the 3D point clouds. However, such images may not be available in many cases, while direct 3D point cloud segmentation methods require semantic class supervision and are challenging to deploy in a zero-shot manner. To address these issues, we present SplatSeg3D, which segments a bare colored point cloud in an open-vocabulary manner without requiring the accompanying images, camera poses, class supervision, or instance mask proposals. To achieve this, we first surfelize the input point cloud and obtain oriented 2D Gaussians. We then render views of the scene by splatting, with a view selection strategy that maximizes scene coverage with as few views as possible. Next, we apply text-prompted segmentation on the rendered views, giving each 3D point a set of multi-view semantic cues, fused to produce a 3D segmentation of the scene. We evaluate SplatSeg3D on semantic segmentation and novel-class instance segmentation on 4 standard benchmarks: Scannet, Scannet++, Replica, and S3DIS. Our pipeline not only performs better or on-par with the state-of-the-art, but we also achieve these results efficiently across runtime and compute requirements. Code will be published.
\end{abstract}

\section{Introduction}
\label{sec:intro}

\begin{figure}[t]
    \centering
    \includegraphics[width=1.0\linewidth]{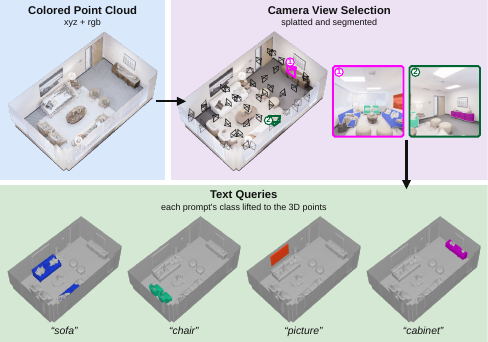}
    \caption{\textbf{\method} segments a bare colored point cloud from text prompts. It surfelizes the point cloud, generates virtual camera views and renders them by
    splatting, then lifts 2D foundational masks back to the 3D points. It requires no training on point cloud data making the pipeline completely training-free.
    }
    \label{fig:teaser}
\end{figure}

Open-vocabulary segmentation of 3D scenes has advanced mostly by leveraging 2D foundational features. Segmentation and
vision-language models (VLMs) trained on billions of images~\cite{sam,clip,sam3} demonstrate strong open-vocabulary segmentation capabilities.
Therefore, most methods in the open-vocabulary 3D segmentation literature perform lifting of multi-view image VLM and/or segmenter outputs onto the 3D point cloud~\cite{openmask3d, jopp3d}. This imposes a strict requirement of posed image sequences or scene panoramas to accompany the point cloud. However, in practice, often only a 3D point cloud might be available, by design or by choice (\eg from LiDAR or terrestrial scanner devices).

On the other hand, methods that can handle point cloud-only data, train a 3D network by supervision on semantic labels~\cite{ptv3, sonata, concerto},
or by supervising a class-agnostic 3D mask proposal module~\cite{openins3d}. 
In this work, we push the open-vocabulary 3D segmentation limits to the case when only a 3D point cloud is available with no accompanying images. 

To tackle this challenge, \method (\cref{fig:teaser}) synthesizes its own 2D image views directly from the input point cloud. However, a naive projection of raw points to get the image renders leads to insufficient quality for accurate image segmentation. Instead we propose surfelization of the point cloud, where at every 3D point we place a surfel~\cite{surfels}, a flat 2D Gaussian disk that is aligned with the local surface normal.
The surfels are computed in closed form and are standard 2D Gaussian splatting primitives~\cite{gs2d}. 

\begin{figure}[t]
    \centering
    \includegraphics[width=1.0\linewidth]{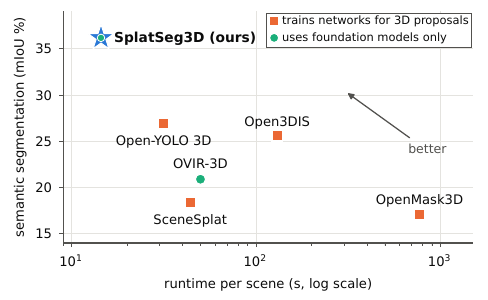}
    \caption{\textbf{Segmentation performance against runtime on Replica.} The plot depicts mIoU over the $51$ classes (\cref{tab:replica})
    against the time to load the models and process one scene (\cref{tab:runtime}). \method demonstrates a better ratio of performance over runtime.
    }
    \label{fig:speed}
\end{figure}

When working with captured images of scans, flexibility in view selection is limited. In our case, a camera placement strategy is used to minimize the number of views while maximizing point cloud coverage, which is the main driver of efficiency in \method. We frame this camera placement as a depth-tested coverage problem over the point cloud, and design an automatic camera pose selector. Segmentation masks of the selected views can then be lifted to the point cloud's 3D points. The final semantic labels of points then become the fused votes from the 3D lifted masks, and
instances are the connected components of each semantic class, requiring no supervised 3D training or instance proposal network. Our contributions are summarized as follows:
\begin{itemize}
    \item {\method}, an open-vocabulary 3D segmentation method whose only input is a colored point cloud, without any images, camera poses, or trained instance proposals. Instead, \method renders its own virtual views through point cloud surfelization and splatting.
    \item {View selection} algorithm for the splatted views, posed as a depth-tested view-coverage problem, minimizing the number of views while maximizing the point cloud coverage.
    \item {Zero-shot semantic and novel-class instance segmentation} on four standard benchmarks, performing on-par with previous open-vocabulary 3D segmentation methods, and exceeding all zero-shot baselines.
    \item {Efficiency in compute and runtime}: \method performs 3D segmentation while requiring much lower GPU compute and runtime than other methods (\cref{fig:speed}). In combination with our zero-shot capability, the method is also more efficient to deploy. 
\end{itemize}

\section{Related Work}
\label{sec:related}

\paragraph{Open-Vocabulary 3D Segmentation.}
The most common open-vocabulary segmentation approaches lift predictions on 2D images accompanying the point clouds into 3D. OpenScene~\cite{openscene}
fuses per-pixel vision-language features into the points of the scene.
OpenMask3D~\cite{openmask3d} queries class-agnostic 3D proposals from a supervised Mask3D~\cite{mask3d}
by aggregating CLIP~\cite{clip} features over the top views of each proposal, and
JOPP-3D~\cite{jopp3d} and Open-YOLO 3D~\cite{openyolo3d} refine the same concept further by either exploiting 3D-to-panorama correspondences or making the 3D segmentation faster, respectively.

A parallel line of open-vocabulary works~\cite{openscene, mosaic3d} trains a 3D network on supervision signals derived from the scan's images (distilled 2D features or captions), so that inference only requires the point cloud. OpenScene's~\cite{openscene} 3D variant
distills 2D features into a convolutional backbone. PLA, RegionPLC, and
Lowis3D~\cite{pla,regionplc,lowis3d} train on captions paired with the 3D points of a benchmark. 
SceneSplat~\cite{scenesplat} differs in that it trains and infers on the same Gaussian scenes reconstructed from captured images. 
On the other hand, there are recent methods that can directly operate on point cloud data with specialized 3D network architectures~\cite{pointins, dos, vino}. 
These recent methods also perform better 
because they extract foundational features directly from 3D data and pair them with a supervised classifier to segment the point cloud, whereas the before mentioned open-vocabulary methods only rely on 2D foundational features.

Zero-shot methods instead lift 2D masks without any 3D training. They merge
masks over frames~\cite{yang2023sam3d, jopp3d},
and fuse 2D detections into 3D instances that a text query retrieves by feature similarity~\cite{ovir3d},
or merge superpoints~\cite{connected} guided by 2D masks~\cite{povo}. GoDeep~\cite{godeep} labels points instead of instances with a VLM that describes each posed image, parses the description into object names, segments them, and lifts the masks to 3D.
\method builds upon these concepts of zero-shot methods, so they are our primary comparison. However, we provide indicative comparisons with other open-vocabulary methods that require dataset-specific training.

\begin{table}
  \centering\footnotesize
  \setlength{\tabcolsep}{1.5pt}
  \caption{\textbf{\method Positioning.} Columns: no images at training or inference; no
  network trained on 3D data, unless only foundation model-based; no instance proposals from supervised backbones.
  }
  \begin{tabular}{@{}lccc@{}}
    \toprule
    \shortstack{Method setting} & \shortstack{Point cloud\\only} & \shortstack{No \\3D training} & \shortstack{No\\proposals} \\
    \midrule
    Posed-image fusion~\cite{openscene}                          & \xmarkc & \cmarkc & \cmarkc\\
    Proposal backbones~\cite{openmask3d,open3dis,openyolo3d}        & \xmarkc & \xmarkc & \xmarkc\\
    2D mask merging~\cite{yang2023sam3d,jopp3d}          & \xmarkc & \cmarkc & \cmarkc\\
    3D networks~\cite{sonata,pointins}                 & \cmarkc & \xmarkc & \cmarkc\\
    Rendered snapshots~\cite{openins3d}                          & \cmarkc & \xmarkc & \xmarkc\\
    \midrule
    \textbf{\method (ours)}                                      & \cmarkc & \cmarkc & \cmarkc\\
    \bottomrule
  \end{tabular}
  \label{tab:taxonomy}
\end{table}

\paragraph{Rendering Point Clouds and Splatting.}
Rendering 3D data to train 2D networks has been explored as a concept. Multi-view CNNs~\cite{mvcnn} classify shapes from rendered
views and
PointCLIP~\cite{pointclip,pointclipv2} projects point clouds to depth maps for zero-shot CLIP.
These approaches operate on single objects, where a camera trajectory around them is sufficient. However, in larger scenes, camera placement requires further consideration, \eg due to self-occlusions. The closest prior work at scene scale is OpenIns3D~\cite{openins3d},
whose snap module renders synthetic scene-level images of the point cloud and a "Lookup" module
assigns a class to 3D mask proposals from 2D detections on them. However, its proposals come from a trained class-agnostic
3D mask module~\cite{mask3d}, and its synthesized images are used to classify them.

On the other hand, Gaussian primitives are usually optimized against images: 
3D Gaussian Splatting~\cite{gs3d} and its 2D variant~\cite{gs2d} fit Gaussian primitives to match target images. 
Segmentation via Gaussian splatting~\cite{scenesplat} is usually performed via 2D semantics on the same target images used for optimization. 
These semantic features are then either optimized alongside the color~\cite{langsplat} or uplifted into an already optimized Gaussian scene~\cite{xie_2026_relags}. 
In \method, segmentation is performed on the rendered views, from which both semantics and instances are built, and no 3D network is trained at any stage.

\paragraph{View Selection \& Planning.}
View planning has been used in active reconstruction~\cite{connolly1985nbv,scott2003viewplanning}, where next-best-view planners
select sensor poses to look for the unknown geometry in the scene.
Our setting inverts that premise - the geometry is fully known, and the observation is given, but the objective is coverage of a scene at
sufficient quality which is what a 2D segmenter can {process}. 

\noindent To conclude this section, \Cref{tab:taxonomy} positions \method among the various 3D segmentation approaches discussed above. Existing methods do not fulfill at least one of the three stated properties. Our approach meets all three properties while performing at state-of-the-art quality.

\section{Method}
\label{sec:method}

\begin{figure*}
    \centering
    \includegraphics[width=1.0\linewidth]{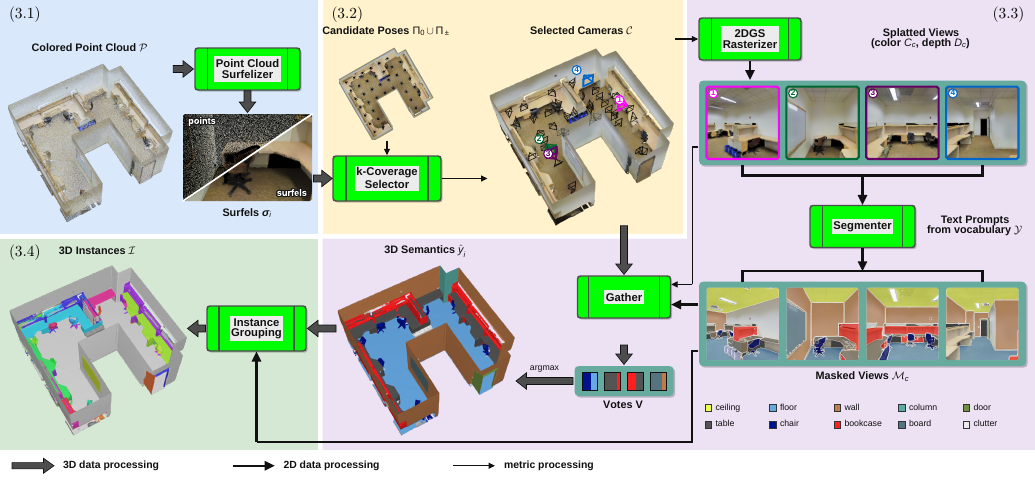}
    \caption{\textbf{\method pipeline}.
    (3.1)~Every point becomes a surfel, oriented by its normal and sized by its
    spacing. The difference between direct 3D point image projection and surfelization is shown. (3.2)~Candidate camera poses stand on a grid, each looking at various yaws, and a second tier is pitched $\pm\phi$. The $k$-coverage
    selector admits only depth-tested views at sufficient resolution and stops on its own after ensuring enough coverage.
    The colored frustums are the cameras of the four views on the right.
    (3.3)~The views are splatted with color and depth and segmented from text prompts.
     Every point reads the masks at its own
    depth-tested pixel, and its score-weighted votes are fused by argmax.
    (3.4)~The connected components of each class are split into instances by the per-view masks. Wide grey arrows indicate 3D data, thick black arrows 2D data, and thin arrows
    camera poses and depth.
    }
    \label{fig:method}
\end{figure*}

Given a colored point cloud $\mathcal{P} = \{(\mathbf{x}_i, \mathbf{c}_i)\}_{i=1}^{N}$ in $(\text{xyz, rgb})$ format, 
and a vocabulary $\mathcal{Y}$ of class names, we predict a semantic label $\hat{y}_i$ for every point and a set of instances $\mathcal{I}$.

\noindent\emph{Notation:} A surfel built from a point $i$ is denoted $\boldsymbol{\sigma}_i = (\mathbf{x}_i, \mathbf{n}_i, r_i, \mathbf{c}_i)$: it is centered at the point $\mathbf{x}_i$, has normal $\mathbf{n}_i$ and radius $r_i$, and carries the point's color $\mathbf{c}_i$. A camera view $c$ maps world to camera coordinates with shared camera intrinsics $\mathbf{K}$ containing focal length $f$ and projects a point $\mathbf{x}$ to pixel $\mathbf{u}_c(\mathbf{x})$ at depth $z_c(\mathbf{x})$. $C_c$, $D_c$, and $A_c$ are the
rendered color, depth, and alpha images of view $c$, defined on the pixels $\mathbf{u} \in \Omega$ of its pixel grid. The rendered depth $D_c(\mathbf{u})$ is that of the surface seen at pixel $\mathbf{u}$. $\mathcal{M}_c$ is the set of masks of view $c$. $\mathbf{V} \in \mathbb{R}_{\ge 0}^{N \times |\mathcal{Y}|}$
holds the per-point votes $V_{iy}$, and $[\cdot]$ is the Iverson bracket. 

For each scene, our proposed method (\cref{fig:method}) performs the following steps:
\begin{enumerate}[label=(\alph*),itemsep=1pt,topsep=2pt]
        \item \textbf{surfelize} every point $i$ of the point cloud $\mathcal{P}$ into a surfel $\boldsymbol{\sigma}_i$ (\cref{sec:disks});
    \item \textbf{generate} candidate cameras $\Pi$ in the scene, and \textbf{select} the camera views $\mathcal{C}$ from them efficiently (\cref{sec:cameras});
    \item \textbf{render} each $c \in \mathcal{C}$ by splatting and \textbf{segment} it into masks
          $\mathcal{M}_c$ from text prompts (\cref{sec:segment});
    \item \textbf{gather} the masks into 3D point votes $\mathbf{V}$ and \textbf{fuse} them into labels $\hat{y}_i$
          (\cref{sec:gather});
    \item \textbf{group} the labelled points into instances $\mathcal{I}$ 
    (\cref{sec:instances}).
\end{enumerate}

\subsection{Point Cloud Surfelization}
\label{sec:disks}

\begin{figure}[t]
    \centering
    \includegraphics[width=1.0\linewidth]{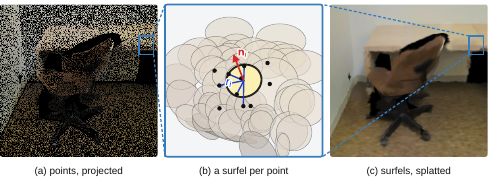}
    \caption{\textbf{Views from points}. (a)~The raw
    points projected into one chosen view, one pixel each.
    (b)~Zoomed patch of the desk's front edge (boxes in a and c). Every point carries a surfel in the plane
    orthogonal to its normal $\mathbf{n}_i$, fitted to its $k_{\mathrm{n}}$ nearest neighbors (dots), with radius
    $r_i$, the mean distance to its $k_{\mathrm{r}}$ nearest (blue). (c)~The surfels splatted by the 2DGS rasterizer is input to the segmenter.}
    \label{fig:disks}
\end{figure}

Direct projection of the point cloud creates sparse images with background artifacts (\cref{fig:disks}) that 2D segmenters fail to segment accurately, therefore, we first perform a surfelization of the point cloud to improve rendered image quality. The
surface normal is chosen as the least-variance direction of its $k_{\mathrm{n}}$ nearest neighbors, oriented toward the point cloud's centroid $\bar{\mathbf{x}}$,
\begin{equation}
  \mathbf{n}_i = \mathrm{eig}_{\min}\big(\mathrm{Cov}(\mathcal{N}_{k_{\mathrm{n}}}(\mathbf{x}_i))\big),
  \qquad \langle \mathbf{n}_i, \bar{\mathbf{x}} - \mathbf{x}_i \rangle \ge 0 ,
  \label{eq:normal}
\end{equation}
and the radius is chosen to be the mean distance to its $k_{\mathrm{r}}$ nearest neighbors (such that neighboring disks overlap and close the gaps between points), clipped to the point cloud's $1$st and
$99$th percentiles,
\begin{equation}
  \begin{aligned}
    \ell_i &= \tfrac{1}{k_{\mathrm{r}}} \textstyle\sum_{j \in \mathcal{N}_{k_{\mathrm{r}}}(\mathbf{x}_i)}
              \|\mathbf{x}_j - \mathbf{x}_i\| , \\
    r_i &= \mathrm{clip}\big(\ell_i, P_1(\ell), P_{99}(\ell)\big) .
  \end{aligned}
  \label{eq:radius}
\end{equation}
Point $i$ becomes the surfel $\boldsymbol{\sigma}_i$, a 2D Gaussian of standard deviation
$r_i$ in the plane through $\mathbf{x}_i$ orthogonal to $\mathbf{n}_i$, with opacity $o = 0.99$ and the point's color $\mathbf{c}_i$. %
The 2DGS rasterizer~\cite{gs2d} performs a sorting of all surfels that project onto the image. Compositing of surfels, that overlap a pixel is performed in front-to-back order, indexed by $j$:
\begin{equation}
  \begin{aligned}
    \alpha_j &= o\, e^{-\|\boldsymbol{\xi}_j - \mathbf{x}_j\|^2 / 2 r_j^2} , \quad
    w_j = \alpha_j \textstyle\prod_{j' < j} (1 - \alpha_{j'}) , \\
    C_c &= \textstyle\sum_j w_j\, \mathbf{c}_j , \quad
    D_c = \textstyle\sum_j w_j\, z_j , \quad A_c = \textstyle\sum_j w_j ,
  \end{aligned}
  \label{eq:render}
\end{equation}
where $\boldsymbol{\xi}_j$ is the point at which the ray meets the plane of $\boldsymbol{\sigma}_j$ and $z_j$ is the
camera depth of $\boldsymbol{\sigma}_j$. Every view with a width of $W$ pixels, has the field of view
$\theta_{\mathrm{fov}}$, so $f = W / (2 \tan(\theta_{\mathrm{fov}} / 2))$.

\subsection{View Selection}
\label{sec:cameras}

\begin{figure*}[t]
    \centering
    \includegraphics[width=\linewidth]{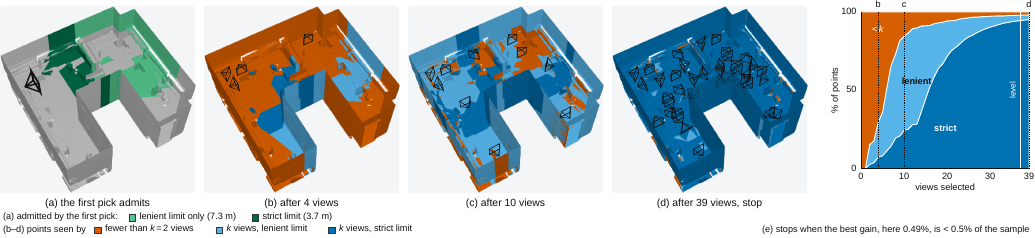}
    \caption{\textbf{Where to look}, camera selection replayed step by step.
    (a)~The first pick, the candidate with the largest gain. 
    (b--d)~Every point's state after
    progressively increasing the camera view picks. Early picks focus on coverage and once a surface is seen by $k=2$ views at
    the lenient limit, later picks focus on its resolution at the strict limit. (e)~The share of points in each state after every pick. The level tier runs to convergence at certain number of  views, 
    and the selector stops when the best remaining gain is $<0.5\%$ of its sample. 
    }
    \label{fig:selection}
\end{figure*}

We generate candidates from the free space of the scene and admit views by their resolution capacity (ground sampling distance, \cref{eq:vis}) to efficiently place the cameras. 

\noindent\textbf{Computing candidate poses.}
We start with cameras standing at eye height above the floor
($\approx1.6$\,m).
A grid of spacing $h$ over the
surface, checked against an occupancy grid of $\delta$ voxel size, gives us the free positions $\mathcal{G}$.
From each position, we generate different views by selecting $n_\psi$ camera yaws:
${\Psi = \{2\pi j / n_\psi : 0 \le j < n_\psi\}}$ at a pitch $\phi$. We only retain camera poses $\Pi_{\phi}$, if at least a fraction $\eta_{\min}$ points ($N_\eta$ of the total points $N$), falls into its image:
\begin{equation}
   \Pi_{\phi} = \big\{ c \in \mathcal{G} \times \Psi \times \{\phi\} : \eta(c) \ge \eta_{\min} \big\} ,
  \label{eq:pool}
\end{equation}
where a candidate $c$ is placed at a position in $\mathcal{G}$, and $\eta(c)$ is the fraction from that candidate. The first tier $\Pi_0$ looks horizontally, and a second tier
$\Pi_{\pm} = \Pi_{+\phi} \cup \Pi_{-\phi}$ pitches up and down by
$\phi$.

\noindent\textbf{Admission by resolution and visibility.}
A point $\mathbf{x}$ lands in view $c$ at pixel $\mathbf{u} = \mathbf{u}_c(\mathbf{x})$ and depth
$z = z_c(\mathbf{x})$. 
The visibility test of points are formed in the following way: 
A view admits a visible point only if its ground sampling distance
$z/f$ (the width of the surface patch that one pixel covers at a point's depth) is below a resolution limit $g$,
\begin{equation}
  \begin{aligned}
    \mathrm{vis}_c(\mathbf{x}) &= [\mathbf{u} \in \Omega] \wedge [A_c(\mathbf{u}) > 0.05]
                                \wedge [z \le 1.05\, D_c(\mathbf{u})] , \\
        v_c^{g}(\mathbf{x}) &= \mathrm{vis}_c(\mathbf{x}) \wedge [z / f \le g] .
  \end{aligned}
  \label{eq:vis}
\end{equation}
The resolution limit $g$ computed this way becomes a generic property of the segmenter and renderer without requiring a scene-specific fit.
We define a lenient limit
$g_{\text{len}}$ and a strict one $g_{\text{str}} (< g_{\text{len}})$.

\noindent\textbf{Coverage-coupled selection.}
For each point $\mathbf{x}$ in a set $\mathcal{Q}$ of surfel centers, the selector counts the
selected views that admit it at the two limits,
$L(\mathbf{x}) = \sum_{c \in \mathcal{C}} v_c^{g_{\text{len}}}(\mathbf{x})$ and
$S(\mathbf{x}) = \sum_{c \in \mathcal{C}} v_c^{g_{\text{str}}}(\mathbf{x})$. A point is {covered}
once $L(\mathbf{x}) \ge k$ and {resolved} once $S(\mathbf{x}) \ge k$, with $k = 2$; these are
the states colored in \cref{fig:selection}. A candidate's gain counts the points that still need what
it offers: (1) the uncovered points it admits at the lenient limit, and (2) the covered but unresolved points it
admits at the strict limit (\cref{alg:kcov}, lines~\ref{alg:qcov}--\ref{alg:gain}). Resolution is thus
sought only where coverage is already met, which couples the two limits. The selector adds the
candidate with the largest gain, updates $L$ and $S$, and repeats, so early picks focus on coverage and
later ones on resolution. It stops on its own once the best gain falls below $0.5\%$ of $|\mathcal{Q}|$.
It runs on the horizontal tier $\Pi_0$
first and then on the pitched tier $\Pi_{\pm}$, starting from the counts the $\Pi_0$ views left.

\begin{algorithm}[t]
  \caption{$k$-Coverage selector}
  \label{alg:kcov}
  \begin{algorithmic}[1]
    \Require tiers $\Pi_0$, $\Pi_{\pm}$ from \cref{eq:pool}; sample $\mathcal{Q}$; coverage $k$;
      admission $v_c^{g}$ from \cref{eq:vis} at the limits $g_{\text{len}} > g_{\text{str}}$
    \Ensure selected views $\mathcal{C}$
    \State render $D_c$ and $A_c$ of every $c \in \Pi_0 \cup \Pi_{\pm}$ \Comment{\cref{eq:render}}
    \State $\mathcal{C} \gets \emptyset$, \ $L \gets 0$, \ $S \gets 0$ \Comment{on all of $\mathcal{Q}$}
    \For{$\Pi \in (\Pi_0, \Pi_{+}, \Pi_{-})$} \Comment{horizontal tier first}
      \Loop
        \State $\mathcal{Q}_{\mathrm{cov}} \gets \{\mathbf{x} \in \mathcal{Q} : L(\mathbf{x}) < k\}$ \label{alg:qcov}
        \State $\mathcal{Q}_{\mathrm{res}} \gets \{\mathbf{x} \in \mathcal{Q} : L(\mathbf{x}) \ge k,\ S(\mathbf{x}) < k\}$
        \ForAll{$c \in \Pi \setminus \mathcal{C}$}
            \State \label{alg:gain} $\begin{aligned}[t]
            \mathrm{gain}(c) \gets {} & \textstyle\sum_{\mathbf{x} \in \mathcal{Q}_{\mathrm{cov}}} v_c^{g_{\text{len}}}(\mathbf{x}) \\
                                     + {} & \textstyle\sum_{\mathbf{x} \in \mathcal{Q}_{\mathrm{res}}} v_c^{g_{\text{str}}}(\mathbf{x})
          \end{aligned}$
        \EndFor
        \State $c^{\star} \gets \argmax_{c \in \Pi \setminus \mathcal{C}} \mathrm{gain}(c)$
        \If{$\mathrm{gain}(c^{\star}) < 0.5\%(\,|\mathcal{Q}|)$}
          \State \textbf{break} \Comment{next tier}
        \EndIf
        \State $\mathcal{C} \gets \mathcal{C} \cup \{c^{\star}\}$
        \State $L \gets L + v_{c^{\star}}^{g_{\text{len}}}$, \ $S \gets S + v_{c^{\star}}^{g_{\text{str}}}$
      \EndLoop
    \EndFor
    \State \Return $\mathcal{C}$
  \end{algorithmic}
\end{algorithm}

\subsection{View Segmentation and Lifting by Gathering}
\label{sec:segment}
\label{sec:gather}
Each selected view is segmented by a 2D open-vocabulary segmenter~\cite{sam3}. For view $c$, the applied class prompts give a set of masks $\mathcal{M}_c$. Each mask is a
triple $m = (M_m, y_m, s_m)$: a binary function $M_m$ that is $1$ on the pixels $\mathbf{u}$ it
covers, the class $y_m$ it was prompted for, and the confidence score $s_m$ from the segmenter.
$\mathcal{M}_c^{y} = \{ m \in \mathcal{M}_c : y_m = y \}$ contains the masks of class $y$.

To setup the vote gathering, the masks are first projected onto the point cloud $\mathcal{P}$. 
Each of the points in $\mathcal{P}$ then looks up which masks
contain its pixel $\mathbf{u}_c(\mathbf{x}_i)$, and every such mask of class $y$ adds its score to the point's vote for $y$:
\begin{equation}
   V_{iy} = \sum_{c \in \mathcal{C}} \sum_{m \in \mathcal{M}_c^{y}} s_m
  \big[\mathrm{vis}_c(\mathbf{x}_i) \wedge M_m(\mathbf{u}_c(\mathbf{x}_i))\big] ,
  \label{eq:vote}
\end{equation}
The above visibility test in \cref{eq:vote} is similar to the test that view selection uses in \cref{eq:vis} of~\cref{sec:cameras}. However, here the mask pixel lookup is used instead of the resolution term, but both filters are based on $\mathrm{vis}_c(\mathbf{x}_i)$.
Each point is then assigned the semantic class label by majority of votes. To avoid discontinuities, a point without votes takes the label of the nearest voted point $j^{\star}$ if it lies within a fixed ratio of three times the median surfel radius $\tilde{r}$,
and otherwise stays unlabelled ($\varnothing$):
\begin{equation}
  \hat{y}_i =
  \begin{cases}
        \argmax_{y} V_{iy} , & \sum_{y} V_{iy} > 0 , \\
    \hat{y}_{j^{\star}} , & \|\mathbf{x}_i - \mathbf{x}_{j^{\star}}\| \le 3\,\tilde{r} , \\
    \varnothing , & \text{otherwise} .
  \end{cases}
  \label{eq:fuse}
\end{equation}

\subsection{Instances from Semantics}
\label{sec:instances}

Instances are constructed from the fused semantic class labels using a modified connected-components approach~\cite{clasicconnected}. For each predicted class, its points are first voxelized on a grid of size $\delta_{\mathrm{I}}$, and the components of occupied voxels connected through faces, edges, or corners~\cite{digitialti} 
are our candidate instances. Connected components cannot separate adjacent objects of one class~\cite{pointgroup}, but the 2D segmenter can.

We therefore split each component of class $y$ by the segmenter's per-view masks $\mathcal{M}_c^{y}$. 
We start in a view $c$, where every point and voxel of the component is assigned a mask. The mask $\nu_c(i)$ of a
point $i$ the view sees is first computed in 2D, as the highest-scoring mask of $\mathcal{M}_c^{y}$ that
contains its pixel. The mask $\mu_c(a)$ of a voxel $a$ is then aggregated in 3D, as the mask that most of its seen points agree on:
\begin{equation}
  \begin{aligned}
    \nu_c(i) &= \argmax_{m \in \mathcal{M}_c^{y} :\, M_m(\mathbf{u}_c(\mathbf{x}_i)) = 1} s_m , \\
    \mu_c(a) &= \operatorname*{mode}_{i \in a :\, \mathrm{vis}_c(\mathbf{x}_i)} \nu_c(i).
  \end{aligned}
  \label{eq:voxmask}
\end{equation}
A link $(a, b)$ between
neighboring voxels then weighs the views that put both ends in one mask against those that put them
in two,
\begin{equation}
  w_{ab} = \sum_{c \in \mathcal{C}_{ab}} \big( 2\,[\mu_c(a) = \mu_c(b)] - 1 \big) ,
  \label{eq:link}
\end{equation}
where $\mathcal{C}_{ab}$ holds the views in which both voxels take a mask. The component's voxels are
then grouped by correlation clustering over its links $\mathcal{E}$,
\begin{equation}
  \kappa^{\star} = \argmax_{\kappa} \sum_{(a,b) \in \mathcal{E}} w_{ab}\, [\kappa(a) = \kappa(b)] ,
  \label{eq:partition}
\end{equation}
over the assignments $\kappa$ of voxels to clusters,
solved by greedy additive edge contraction (GAEC)~\cite{keuper2015multicut}, which merges the two adjacent
clusters with the largest positive summed weight.
Smaller clusters join their largest neighbor, or are dropped if
they are adjacent to none. Each remaining cluster $q$ is an instance of class $y_{\mathcal{I}} = y$,
$\mathcal{I} = \{ i : \hat{y}_i = y,\ \kappa^{\star}(a_i) = q \}$, with $a_i$ the voxel of point $i$.

\section{Experiments}
\label{sec:exp}

\subsection{Implementation Details}
\label{sec:impl}

Across all datasets we use the same constant values mentioned in \cref{sec:method}: 
$k_{\mathrm{n}} = 12$
and $k_{\mathrm{r}} = 6$ neighbors, and views are $W=512$\,px with
$\theta_{\mathrm{fov}} = 70^\circ$ ($f = 365.6$\,px). 
$h = 1$\,m, $\delta = 10$\,cm, $n_\psi = 8$ yaws and $\phi = 20^\circ$. $\eta_{\min} = 5\%$ of a sample of $N_\eta = 10{,}000$ points. The resolution limits are $g_{\text{str}} = 0.010$\,m/px and $g_{\text{len}} = 0.020$\,m/px
, and $|\mathcal{Q}| = 20{,}000$ surfel centers.
Instances are built on voxels of $\delta_{\mathrm{I}} = 5$\,cm.
{No camera, rendering or fusion constant is changed for any benchmark}. 
All experiments ran on one RTX A6000.

\begin{table}
  \centering\small
  \setlength{\tabcolsep}{5pt}
  \caption{\textbf{Point cloud semantic segmentation on Replica}, under OpenNeRF's
  protocol~\cite{openerf}.
  Columns: \emph{Proposal train}: the 3D data on which a method trains its proposals or its 3D network, \emph{none} meaning that only frozen foundation models are used. \emph{Modality}: the data a method uses, in training or at inference.
  $^\dagger$Run by us with its released code.
  $^{\ast}$ScanNet, ScanNet++, and Matterport3D. In each group, best result is \textbf{bold}, and best overall is also \underline{\textbf{underlined}}.
  \method performs better than other open-vocabulary methods on this evaluation-only dataset.
   }
  \resizebox{\linewidth}{!}{
  \begin{tabular}{llccc}
    \toprule
    \textbf{Method} & Modality & Proposal train & mIoU(\%) & mAcc(\%) \\
    \midrule
    OpenScene-3D~\cite{openscene}       & 2D+3D & ScanNet    & 15.9 & 24.6 \\
    OpenMask3D$^\dagger$~\cite{openmask3d} & 2D+3D & ScanNet200 & 17.1 & 29.6 \\
    SceneSplat$^\dagger$~\cite{scenesplat} & 2D+3D & 3 datasets$^{\ast}$ & 18.4 & 31.6 \\
    Open3DIS~\cite{open3dis}              & 2D+3D & ScanNet200 & 25.6 & 38.7 \\
    OVO~\cite{ovo}                        & 2D+3D & ScanNet++  & 26.5 & 35.8 \\
    Open-YOLO 3D$^\dagger$~\cite{openyolo3d} & 2D+3D & ScanNet200 & 26.9 & 35.3 \\
    Ov3R~\cite{ov3r}                      & 2D+3D & ScanNet++  & \textbf{31.9} & \textbf{42.3} \\
    \midrule
    OVIR-3D$^\dagger$~\cite{ovir3d}       & 2D+3D & none & 20.9 & 28.3 \\
    HOV-SG~\cite{hovsg}                   & 2D+3D & none & 22.5 & 34.2 \\
    OVI-MAP~\cite{ovimap}                 & 2D+3D & none & 26.5 & 32.2 \\
    \ours                                 & 3D    & none & \underline{\textbf{36.2}} & \underline{\textbf{47.2}} \\
    \bottomrule
  \end{tabular}}
  \label{tab:replica}
\end{table}

\begin{table}
  \centering\small
  \setlength{\tabcolsep}{5pt}
  \caption{\textbf{Point cloud semantic segmentation on ScanNet++}, validation scenes, under Mosaic3D's protocol~\cite{mosaic3d}.
   $^{\ast}$ScanNet, ARKitScenes, Matterport3D, Structured3D, ScanNet++. $^{\ddagger}$ScanNet, ScanNet++,
  Matterport3D. $^{\dagger}$Evaluated using its released code and scored per mesh vertex under the same protocol as other methods. $^{\S}$As published, scored on its own Gaussian reconstructions.
  Therefore, considering the same point-cloud evaluation protocol \method still leads in this table even though we are zero-shot.
  }
  \resizebox{\linewidth}{!}{
  \begin{tabular}{llccc}
    \toprule
    \textbf{Method} & Modality & Proposal train & mIoU(\%) & mAcc(\%) \\
    \midrule
    OpenScene-3D~\cite{openscene}   & 2D+3D & ScanNet    & 8.8  & 14.7 \\
    RegionPLC~\cite{regionplc}      & 2D+3D & ScanNet    & 11.3 & 20.1 \\
    SceneSplat~\cite{scenesplat}    & 2D+3D & ScanNet    & 14.7 & 24.7 \\
    Mosaic3D~\cite{mosaic3d}        & 2D+3D & ScanNet    & 16.2 & 27.1 \\
    Mosaic3D~\cite{mosaic3d}        & 2D+3D & 5 datasets$^{\ast}$ & 18.0 & 29.0 \\
    SceneSplat$^{\dagger}$~\cite{scenesplat} & 2D+3D & 3 datasets$^{\ddagger}$ & 26.4 & 41.4 \\
    SceneSplat$^{\S}$~\cite{scenesplat}    & 2D+3D & 3 datasets$^{\ddagger}$ & \underline{\textbf{28.4}} & \underline{\textbf{50.0}} \\
    \midrule
    GoDeep~\cite{godeep}            & 2D+3D & none & 16.8 & 39.1 \\
    \ours                           & 3D    & none & \textbf{27.3} & \textbf{43.7} \\
    \bottomrule
  \end{tabular}}
  \label{tab:scannetpp}
\end{table}

\begin{table*}
  \centering\footnotesize
  \caption{\textbf{Novel-class 3D instance segmentation} on the base/novel partitions of PLA~\cite{pla}:
  on (a) S3DIS and (b) ScanNet v2. \emph{No base-train} marks methods that never train on the base split
  with semantic labels. $^{\dagger}$Run by us, implemented similar to JOPP-3D(f)~\cite{jopp3d}. Even though other methods use a dedicated proposal network like Mask3D, \method leads most of them and is on par with the best in specific splits.
  }
  \begin{minipage}{0.48\linewidth}
    \centering
    \setlength{\tabcolsep}{3pt}
  \begin{tabular}{llcccc}
    \toprule
    \multicolumn{6}{c}{\textbf{(a) S3DIS~\cite{s3dis} Area 5}} \\
    \midrule
    \multirow{2}{*}{\textbf{Method}} & \multirow{2}{*}{Modality} & \multirow{2}{*}{\shortstack{No\\base-train}} & \multirow{2}{*}{\shortstack{Proposal\\train}} & \multicolumn{2}{c}{AP$_{50}^{\text{novel}}$(\%)} \\
    \cmidrule(lr){5-6}
    & & & & B8/N4 & B6/N6 \\
    \midrule
    Open3DIS~\cite{open3dis}           & 2D+3D & \xmarkc & S3DIS & 26.3 & 29.0 \\
    OpenIns3D~\cite{openins3d}         & 3D    & \cmarkc & S3DIS & 37.0 & 33.0 \\
    JOPP-3D~\cite{jopp3d}              & 2D+3D & \cmarkc & S3DIS & \underline{\textbf{51.7}} & \underline{\textbf{60.7}} \\
    \midrule
    PoVo~\cite{povo}                   & 2D+3D & \cmarkc & none & 29.1 & 33.4 \\
    JOPP-3D(f)~\cite{jopp3d}           & 2D+3D & \cmarkc & none & 38.1 & 41.9 \\
    \ours                              & 3D    & \cmarkc & none & \textbf{42.5} & \textbf{59.6} \\
    \bottomrule
  \end{tabular}
  \end{minipage}\hfill
  \begin{minipage}{0.48\linewidth}
    \centering
    \setlength{\tabcolsep}{3pt}
    \begin{tabular}{llcccc}
      \toprule
      \multicolumn{6}{c}{\textbf{(b) ScanNet v2~\cite{scannet}, validation scenes}} \\
      \midrule
      \multirow{2}{*}{\textbf{Method}} & \multirow{2}{*}{Modality} & \multirow{2}{*}{\shortstack{No\\base-train}} & \multirow{2}{*}{\shortstack{Proposal\\train}} & \multicolumn{2}{c}{AP$_{50}^{\text{novel}}$(\%)} \\
      \cmidrule(lr){5-6}
      & & & & B8/N9 & B10/N7 \\
      \midrule
      PLA~\cite{pla}                     & 2D+3D & \xmarkc & ScanNet & 25.1 & 21.9 \\
      RegionPLC~\cite{regionplc}         & 2D+3D & \xmarkc & ScanNet & 32.2 & \textbf{32.3} \\
      Lowis3D~\cite{lowis3d}             & 2D+3D & \xmarkc & ScanNet & \underline{\textbf{38.1}} & 31.2 \\
      OpenIns3D~\cite{openins3d}         & 3D    & \cmarkc & ScanNet & 19.5 & 27.9 \\
      \midrule
      SAM3D~\cite{yang2023sam3d}+CLIP$^{\dagger}$~\cite{clip} & 2D+3D & \cmarkc & none & 23.2 & 21.5 \\
      \ours                              & 3D    & \cmarkc & none & \textbf{23.9} & \underline{\textbf{32.5}} \\
      \bottomrule
    \end{tabular}
  \end{minipage}
  \label{tab:novel}
\end{table*}

\subsection{Benchmarks and Protocols}
\label{sec:protocol}

\begin{table*}
  \centering\footnotesize
  \setlength{\tabcolsep}{5pt}
    \caption{\textbf{Efficiency on Replica}.
  \emph{Model load}:
  loading the models onto the GPU, once. \emph{Per scene}: mean over the $8$ Replica scenes. \emph{Total}:
  mean for model load and processing one scene. 
  \emph{Peak GPU mem.}: the most GPU
  memory allocated at once while processing a scene.
Compared to other methods, ours is faster to load \& run and is computationally efficient while performing better.
  }
  \begin{tabular}{llrrrrrr}
    \toprule
    \textbf{Method} & Proposal train & Model load (s) & Per scene (s) & Total (s) & FLOPs per scene ($\times10^{13}$) & Peak GPU mem.\ (GB) & mIoU (\%) \\
    \midrule
    OpenMask3D~\cite{openmask3d}          & ScanNet200 & 13.7 & 755.7 & 769.4 & 417 & 9.6 & 17.1 \\
    Open3DIS~\cite{open3dis}              & ScanNet200 & 16.6 & 113.5 & 130.1 & 172 & \textbf{8.1} & 25.6 \\
    SceneSplat~\cite{scenesplat} & 3 datasets$^{\ast}$ & \underline{\textbf{2.8}} & 41.1 & 43.9 & 39.5 & 21.6 & 18.4 \\
    Open-YOLO 3D~\cite{openyolo3d}        & ScanNet200 & 23.5 & \underline{\textbf{8.0}} & \textbf{31.5} & \underline{\textbf{5.5}} & 17.3 & \textbf{26.9} \\
    \midrule
    OVIR-3D~\cite{ovir3d}                 & none & 13.1 & 36.8 & 49.9 & 16.9 & 22.0 & 20.9 \\
    \ours                                 & none & \textbf{5.7} & \textbf{8.7} & \underline{\textbf{14.4}} & \textbf{6.3} & \underline{\textbf{2.3}} & \underline{\textbf{36.2}} \\
    \bottomrule
  \end{tabular}
  \label{tab:runtime}
\end{table*}

We evaluate \method in semantic segmentation and novel-class instance segmentation settings. We follow the standard retrieval setting~\cite{openmask3d,openyolo3d}: the vocabulary is the benchmark's list of class names, and each class is prompted with its name.

For \textit{ScanNet++}~\cite{scannetpp}, we evaluate on its validation scenes, on the vertices of their released meshes. Semantics follow Mosaic3D's~\cite{mosaic3d}
foreground protocol from its official release.
\textit{Replica}~\cite{replica} is an evaluation-only dataset, evaluated on
OpenNeRF's~\cite{openerf} $51$-class semantic protocol.

\textit{S3DIS}~\cite{s3dis} is evaluated on Area 5. We follow PLA's~\cite{pla} Novel Instance segmentation protocol which reports AP on the base/novel partitions (B8/N4 and B6/N6), over the $12$ classes in the dataset.
\textit{ScanNet v2}~\cite{scannet} is evaluated on its validation scenes, also following PLA's
B8/N9 and B10/N7 partitions.

\subsection{Benchmark Evaluation}
\label{sec:sota}
\label{sec:novel}
\noindent\textit{Semantic segmentation.}
On Replica (\cref{tab:replica}), where all methods are tested in zero-shot, \method leads every method by a clear margin, including Ov3R~\cite{ov3r}, which segments with a
module trained on ScanNet++ and lifts its labels onto ground-truth geometry.
On ScanNet++ (\cref{tab:scannetpp}), \method outperforms every method under the same protocol of scoring on the dataset's own mesh, including SceneSplat~\cite{scenesplat}, which trains on ScanNet++'s own scans. SceneSplat's published results are higher because it uses a different protocol of scoring on the Gaussians of its own reconstructions instead. Even methods trained on ScanNet stay well below, as does Mosaic3D's model trained on five datasets that include ScanNet++. Among methods that are zero-shot, \method outperforms GoDeep~\cite{godeep} by a wide margin, without requiring input images.  

\begin{figure*}
    \centering
    \includegraphics[width=0.8\linewidth]{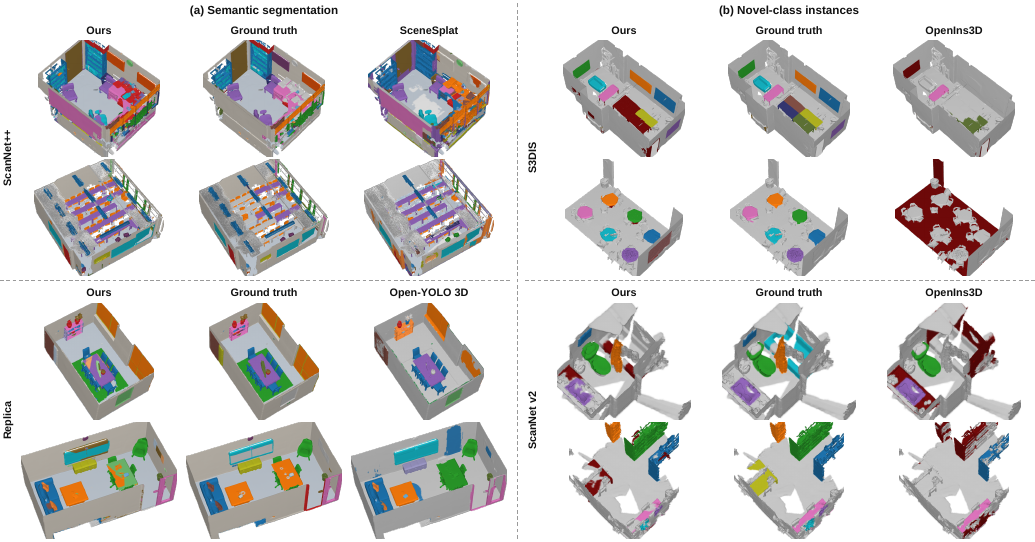}
    \caption{\textbf{Qualitative results}, two scenes per dataset, each against a baseline method from the
    dataset's table that we ran on the same scenes. (a)~Semantic segmentation on ScanNet++ (top-$100$
    classes) against SceneSplat, and on Replica ($51$ classes) against Open-YOLO 3D, whose unlabelled
    points remain grey. 
    (b)~Novel-class
    instances on S3DIS (B6/N6) and ScanNet v2 (B8/N9) against OpenIns3D.
    A ground-truth instance of a novel class and the prediction matching it 
    share one color, unmatched novel predictions are dark red, and base classes are grey. 
    }
    \label{fig:qualitative}
\end{figure*}

\noindent\textit{Novel-class instances.}
\method performs zero-shot segmentation, meaning that every class of a base/novel partition is novel to it. On S3DIS
(\cref{tab:novel}a), it is the highest performing method without a trained instance model on both partitions,
ahead of PoVo~\cite{povo} and JOPP-3D(f)~\cite{jopp3d}, which lift 2D masks from posed images or
panoramas. It is also well ahead of Open3DIS~\cite{open3dis}, which trains on the base classes. Only
JOPP-3D's Mask3D proposals, trained on S3DIS, are higher.
On ScanNet v2 (\cref{tab:novel}b), \method leads every method without base training. Only on B8/N9, the methods that are trained on the base classes (PLA, RegionPLC, and
Lowis3D) stay ahead. On B10/N7, \method leads overall.
\Cref{fig:qualitative} shows qualitative examples of both tasks.

\begin{figure}
    \centering
    \captionof{table}{\textbf{Selector and renderer variants}.}
    {\footnotesize\setlength{\tabcolsep}{3pt}
    \resizebox{0.78\linewidth}{!}{\begin{tabular}{@{}l@{\hspace{9pt}}lccc@{}}
      \toprule
                                     & \tmark{abl-left}variant & pool & views & mIoU\tmark{abl-right} \\
      \midrule
      \textit{selector}              & fixed budget of $64$  & 95.5  & 48.1 & \est{26.3} \\
                                     & adaptive, $k = 1$     & 95.5  & 16.2 & \est{25.0} \\
                                     & \tmark{abl-k2}adaptive, $k = 2$ & 95.5 & 30.2 & \est{26.0} \\
                                     & adaptive, $k = 3$     & 95.5  & 40.3 & \est{25.8} \\
      \cmidrule{1-5}
      \tmark{abl-poses}\textit{poses} & $+$ height from occupancy & 283.9 & 37.0 & \est{26.6} \\
                                     & $+$ pitch $\pm 20^\circ$, one pass & 287.0 & 34.8 & \est{26.7} \\
                                     & \tmark{abl-tiered}$+$ pitch $\pm 20^\circ$, tiered & 287.0 & 40.7 & \textbf{\est{27.3}} \\
                                     & horizontal (\ie $\Pi_0$) & 95.5 & 39.3 & \est{25.5} \\
      \cmidrule{1-5}
      \tmark{abl-renderer}\textit{renderer} & points, z-buffered & 287.0 & 40.7 & \est{22.9} \\
                                     & 3D Gaussians (spheres) & 287.0 & 40.7 & \est{23.7} \\
                                     & surfels, one global radius & 287.0 & 40.7 & \est{25.9} \\
                                     & \tmark{abl-surfels}surfels        & 287.0 & 40.7 & \textbf{\est{27.3}} \\
      \bottomrule
    \end{tabular}\begin{tikzpicture}[remember picture,overlay,
        carry/.style={-{Stealth[length=3.6pt,width=3pt]}, draw=black!55, line width=0.55pt,
                      rounded corners=2pt}]
      \foreach \r in {abl-k2,abl-tiered,abl-surfels}
        \fill[black!8, blend mode=multiply, rounded corners=1.5pt]
          ($(abl-left |- \r) + (-2pt,-2.5pt)$) rectangle ($(abl-right |- \r) + (2pt,6.6pt)$);
      \draw[carry] ($(abl-left |- abl-k2) + (-2pt,2.1pt)$) -| ($(abl-poses) + (6pt,5.2pt)$);
      \draw[carry] ($(abl-left |- abl-tiered) + (-2pt,2.1pt)$) -| ($(abl-renderer) + (6pt,5.2pt)$);
    \end{tikzpicture}\par}
                \vspace{-0.2cm}

    }
    \label{tab:cameras}
    \includegraphics[width=0.7\linewidth]{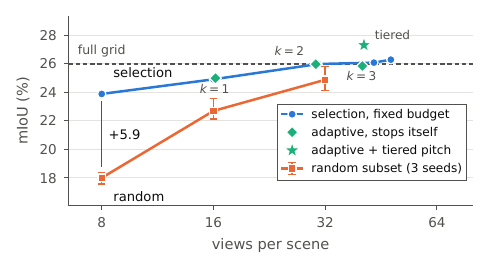}
    \captionof{figure}{\textbf{Choosing the cameras}.}
    \label{fig:frontier}
        \vspace{2pt}
    \begin{minipage}{\linewidth}\small \textbf{Ablations on ScanNet++.}
    Plot: Fixed-budget view selection stops once no view adds a
    new point, so budgets of $32$, $48$ and $64$ average $30.6$, $43.3$ and $48.1$ views. 
    The $k$-adaptive selection 
    instead 
    stops once a view adds $<0.5\%$ of the points. 
    Table: \emph{views} are selected out of a \emph{pool} of candidate poses (avg. over scenes). Overall, the ablations validate our choices in each part of the pipeline.
    \end{minipage}
\end{figure}

\subsection{Efficiency}
\label{sec:runtime}

\Cref{tab:runtime} compares time, computations and memory on Replica, with every method tested on the same
hardware setup. \method takes $14.4$\,s to load and segment per scene on average.
Open-YOLO 3D~\cite{openyolo3d} loads Mask3D~\cite{mask3d} \&
YOLO-World~\cite{yoloworld} and segments in $31.5$\,s, reading the scene's $200$ posed RGB-D frames
and running pre-trained Mask3D for instance proposals.
OVIR-3D~\cite{ovir3d} takes $49.9$\,s, running Detic~\cite{detic}
on every frame, Open3DIS~\cite{open3dis} $130.1$\,s, grounding the text in every frame with Grounding
DINO~\cite{groundingdino}. 

The compute cost depends mainly on the input each method uses and the model applied on them. We run the segmenter, SAM3~\cite{sam3} on the rendered views at $6.3\times10^{13}$ FLOPs per scene. Open-YOLO 3D runs a convolutional detector~\cite{yoloworld} instead on the scene's frames and needs $13\%$ fewer. OVIR-3D needs $2.7$ times FLOPS as ours, SceneSplat $6.3$ times, and Open3DIS and OpenMask3D, which run SAM's ViT-H~\cite{vit} encoder on every frame, are very compute heavy.
\method has the lowest memory usage as well. 
Ours peaks at $2.3$\,GB, averaged per scene, $3.6$ times less than the next, Open3DIS, and $7.6$ times less than Open-YOLO 3D.

\subsection{Ablations}
\label{sec:ablation}

We perform our ablation study on ScanNet++, interchanging one component of our method with standard choices in the pipeline each time.
The full configuration is the main \method pipeline of \cref{tab:scannetpp}.

\noindent\textit{View selection.}
\Cref{fig:frontier} ablates view selection strategy over random selection. At fixed budgets of $8$, $16$ and $32$ views,
selection beats every random subset of the same candidate grid, by a margin that shrinks as the
budget grows. 
\Cref{tab:cameras} further ablates the view selection parameters. "One pass" runs the selector once over all candidates,
while \emph{tiered} runs it on the horizontal poses to convergence and lets the pitched ones only fill
what is remaining in the candidate pool (\cref{alg:kcov}).
Pitched views help when they are tiered behind
the horizontal ones: the tiered selector outperforms the horizontal one even though both converge around the same candidate views.
Sampling camera height from the occupancy column helps less than tiered pitch from a pool
of similar size. 

\noindent\textit{Rendering for segmentation.}
The renderer rows of \cref{tab:cameras} keep our cameras and render them with various primitives.
Projecting the raw points loses the most and
splatting spheres instead of surfels is not much better.
Giving every surfel one global radius (the median
nearest-neighbor spacing) instead of its own, still performs better than using points and spheres.

\subsection{Limitations}
\label{sec:limitations}

\textit{Overall performance is bounded by the segmenter.} Classes that the segmenter fails on, for \eg refrigerator or bathtub, remain undetected at any camera configuration. In some cases, fine-grained
classes are classified as their coarser counterparts due to higher segmenter confidence.

\textit{Instance boundaries remain challenging.} Correlation clustering on
the segmenter's masks separates many nearby objects, but labels still leak onto these
objects. Methods that outperform \method, use supervised instance predictors like Mask3D that have higher fidelity in object boundaries.

\textit{The geometry assumptions are mainly valid for indoor scenes.} Normals are oriented toward the centroid, which
assumes a scene is observed from within, and cameras stand at a fixed eye height above a detected floor, which assumes indoor scenes.

\section{Conclusion}
\label{sec:conclusion}
We introduced \method, an open-vocabulary 3D segmentation pipeline whose only input is a colored
point cloud. It surfelizes the point cloud and chooses camera views to splat, followed by accumulating a frozen text-prompted segmenter's masks back to the points. Because no part of it is trained on 3D labels, every queried class is novel, making the method fully zero-shot.
Through experiments on four benchmarks, we showed that \method outperforms all other zero-shot methods and is on par with open-vocabulary methods that use 3D training or proposals.
In terms of efficiency, \method is deployment-ready with low runtime and compute requirements.
The current mentioned limitations also indicate future research directions such as extending camera placement beyond indoor scans. 

Overall, \method enables open-vocabulary 3D segmentation from point cloud-only data and can potentially inherit future
improvements in foundational 2D segmenters.

\section*{Acknowledgements}
This research was funded by the European Union as part of the project: ShieldBOT (Grant Agreement 101235093).

{
    \small
    \bibliographystyle{ieeenat_fullname}
    \bibliography{main}
}

\clearpage
\renewcommand{\theHequation}{supp.\arabic{equation}}
\renewcommand{\theHtable}{supp.\arabic{table}}
\renewcommand{\theHfigure}{supp.\arabic{figure}}
\appendix

\setcounter{table}{0}\renewcommand{\thetable}{S\arabic{table}}
\setcounter{figure}{0}\renewcommand{\thefigure}{S\arabic{figure}}
\setcounter{equation}{0}\renewcommand{\theequation}{S\arabic{equation}}

\section{More Implementation Details}
\label{sec:supp_impl}

\paragraph{Views per scene.}
Due to the selector algorithm's termination condition (\cref{alg:kcov}), the number of views increases with the size of the scene. It picks
$37.2$ views per Replica~\cite{replica} scene, $34.9$ per S3DIS~\cite{s3dis} room and $42.0$ per ScanNet++~\cite{scannetpp} scene on average. On
ScanNet++ the count ranges from $20$ to $85$ and follows the scene's number of points.

\paragraph{Mask scores and thresholds.}
The segmenter SAM3~\cite{sam3}, encodes each view once and decodes every class prompt against the cached image
features. For class $y$ in view $c$, each of its $200$ query slots returns a detection $m$ with binary
mask $M_m$, class $y_m = y$ and class logit $z_m$, and the slots take the prompt's \emph{presence} logit
$z^{\mathrm{p}}_{y}$ in that view. Of the detections $\mathcal{D}_c$ of all classes, the view filters
those of each class $y$ that clear its threshold $\tau_y$, with $\sigma$ the logistic function,
\begin{equation}
  \begin{aligned}
    s_m &= \sigma(z_m)\,\sigma(z^{\mathrm{p}}_{y}) , \\
    \mathcal{M}_c^{y} &= \{ m \in \mathcal{D}_c : y_m = y,\ s_m > \tau_y \} ,
  \end{aligned}
  \label{eq:score}
\end{equation}
Deciding $\tau_y$: Classes are decoded independently, so masks of different classes may overlap in a view, and fusion
across views (\cref{eq:fuse}) resolves the overlap. A class $y$ is thresholded by $\tau_y = 0.5$ unless that class is rarely proposed (with a rarity factor $\rho_y$). We compute $\rho_y$ as follows,
\begin{equation}
  \begin{aligned}
    \rho_y &= \frac{1}{|\bar{\mathcal{C}}|} \textstyle\sum_{c \in \bar{\mathcal{C}}}
              \big|\{ m \in \mathcal{D}_c : y_m = y,\ s_m > 0.5 \}\big| , \\
    \tau_y &= 0.3 \ \text{ if } \rho_y < 0.02 , \ \text{ else } 0.5 ,
  \end{aligned}
  \label{eq:threshold}
\end{equation}
$\bar{\mathcal{C}}$ is the union of the selected views $\mathcal{C}$ for all scenes in the test set, so each class gets a single threshold $\rho_y$. Without~\cref{eq:threshold} (\ie $\tau_y = 0.5$), performance changes only slightly: mIoU/mAcc is $35.8 / 46.9$ instead of $36.2 / 47.2$. 

\paragraph{Instance confidence.}
Instance prediction evaluation ranks predicted instances by their confidence before computing average precision. Since we use a connected components approach, the instance confidences are computed rather than learning them completely based on a trained network. An
instance $\mathcal{I}$ of class $y_{\mathcal{I}}$ (\cref{sec:instances}) takes its view support, the number of views in which a single mask of its class covers at least half of the points the view sees,
\begin{equation}
  \mathrm{conf}(\mathcal{I}) = \sum_{c \in \mathcal{C}} \max_{m \in \mathcal{M}_c^{y_{\mathcal{I}}}}
  \big[\, |\mathcal{I}_c^{m}| \ge \tfrac{1}{2}\,|\mathcal{I}_c| > 0 \,\big] ,
  \label{eq:supp_conf}
\end{equation}
with $\mathcal{I}_c = \{ i \in \mathcal{I} : \mathrm{vis}_c(\mathbf{x}_i) \}$ the points of
$\mathcal{I}$ that view $c$ sees (\cref{eq:vis}) and
$\mathcal{I}_c^{m} = \{ i \in \mathcal{I}_c : M_m(\mathbf{u}_c(\mathbf{x}_i)) = 1 \}$ those whose
pixel lies in mask $m$. 

\section{Details of Evaluation Protocol}
\label{sec:supp_protocol}

\paragraph{Baselines run by us} ($^{\dagger}$ in \cref{tab:replica,tab:scannetpp,tab:novel}).
On Replica, we run OpenMask3D~\cite{openmask3d}, Open-YOLO 3D~\cite{openyolo3d}, and OVIR-3D~\cite{ovir3d} with
their released code using the Replica configurations on the same $200$ posed RGB-D frames per scene.
For SceneSplat~\cite{scenesplat} on Replica we run its released checkpoint, trained on ScanNet, ScanNet++ and
Matterport3D, on SceneSplat-7K's Gaussian reconstructions of each scene, prompt each class as its
zero-shot tester does and label every vertex by its vote over the $25$ nearest Gaussians. On ScanNet++ the same predictions are scored by Mosaic3D's evaluator like every other reported method in \cref{tab:scannetpp}. 

\paragraph{Details on the SAM3D+CLIP baseline.}
In~\cref{tab:novel}(b), we run a baseline based on only foundation models to compare with \method on ScanNet v2~\cite{scannet}. This baseline is implemented similar to the procedure mentioned in JOPP-3D(f)~\cite{jopp3d}, where SAM3D~\cite{yang2023sam3d} predicted 3D instance masks are aligned with CLIP~\cite{clip} embeddings using the alignment strategy of OpenMask3D~\cite{openmask3d}:

With SAM3D we first lift the 2D masks predicted by SAM~\cite{sam} into 3D instances on the mesh vertices with $5$\,cm voxels. Then, following JOPP-3D(f), we align a VLM embedding to each 3D instance with its top-$5$ views using CLIP. 

\section{Efficiency Evalutation Details}
\label{sec:supp_runtime}

\paragraph{Timing protocol.}
Every method of \cref{tab:runtime} runs on one RTX A6000 over the $8$ Replica scenes, one scene at a
time on an otherwise idle GPU, with its released Replica configuration. For the methods that require image captures, the same $200$ posed RGB-D frames per scene are used. Model load is timed once, and per-scene times are
means over the $8$ scenes after an untimed warm-up scene. SceneSplat's~\cite{scenesplat} time excludes the Gaussian reconstruction its input needs, about $27$ GPU-minutes per
scene on an L4 for SceneSplat-7K. FLOPs count the networks' matrix multiplications,
convolutions and attention from PyTorch's operator counts, with sparse 3D convolutions counted from
their kernel maps. Memory is PyTorch's allocator peak, models included.

\paragraph{Our time per stage.}
On Replica, a scene takes $0.5$\,s for surfelization, $5.4$\,s for camera selection, $0.5$\,s for
rendering, $1.5$\,s for the segmenter, $0.1$\,s for lifting and $0.6$\,s for fusion, $8.7$\,s in total, averaged per scene.

\begin{figure*}[t]
  \centering
  \includegraphics[width=\linewidth]{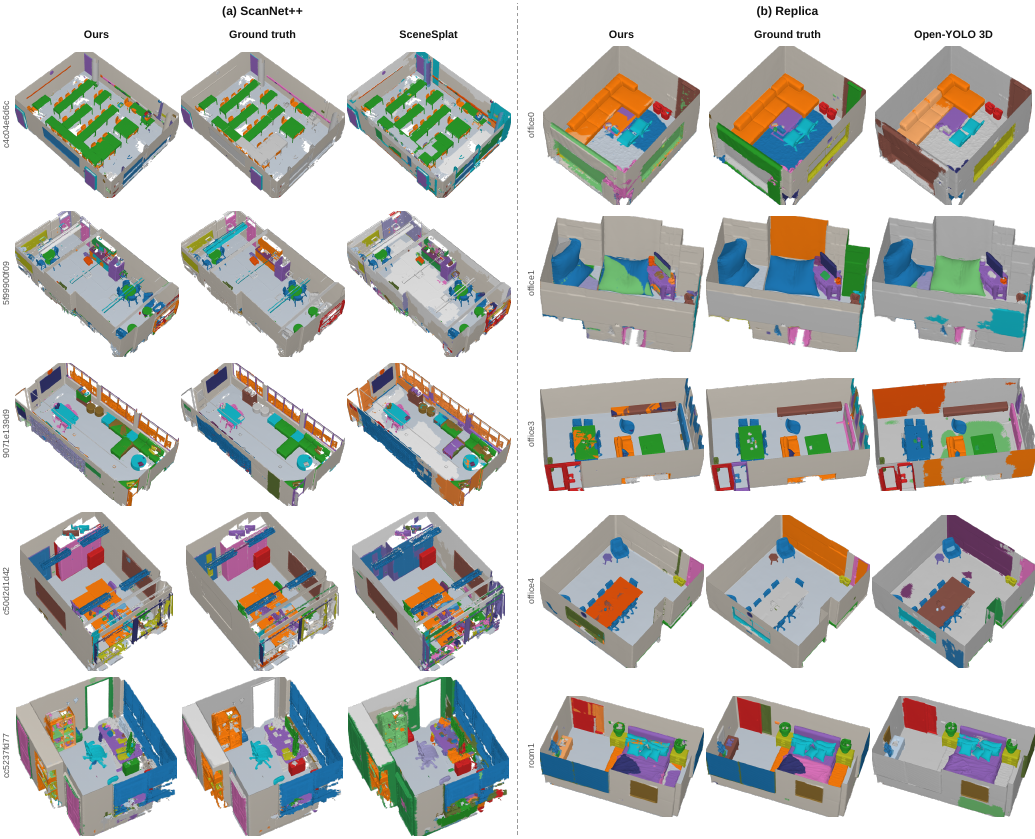}
  \caption{\textbf{Semantic segmentation}, five scenes per dataset: (a)~ScanNet++ (top-$100$ classes)
  against SceneSplat, and (b)~Replica ($51$ classes) against Open-YOLO 3D. Semantic class colors follow the same palette as~\cref{fig:qualitative}, and ceilings are hidden for visualization.}
  \label{fig:supp_qual_sem}
\end{figure*}
\begin{figure*}[t]
  \centering
  \includegraphics[width=\linewidth]{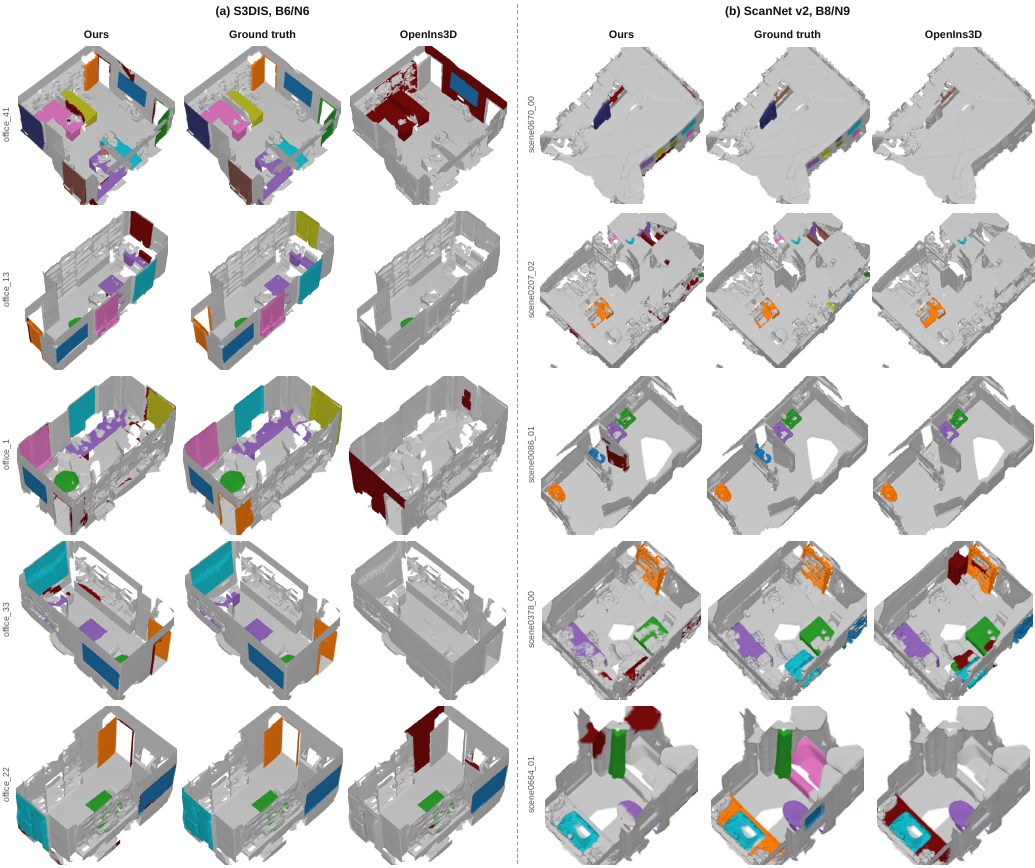}
  \caption{\textbf{Novel-class instances}, five scenes per dataset: (a)~S3DIS (B6/N6) and (b)~ScanNet v2
  (B8/N9), against OpenIns3D~\cite{openins3d}. A ground-truth instance of a novel class and the prediction matching it at
  IoU~$\ge 0.5$ share one color, unmatched novel predictions are dark red, and base classes are grey;
  floor, novel in B6/N6, is left grey, and ceilings are hidden for visualization.}
  \label{fig:supp_qual_inst}
\end{figure*}
\section{Additional Qualitative Results}
\label{sec:supp_qual}

\Cref{fig:supp_qual_sem,fig:supp_qual_inst} show five more scenes per dataset in the format of
\cref{fig:qualitative}. Each dataset is compared to the same
method as in \cref{fig:qualitative}.

\clearpage
\clearpage

\end{document}